\documentclass{ieeeaccess}
\usepackage{amsmath,amssymb,amsfonts}
\usepackage{algorithmic}
\usepackage{graphicx}
\usepackage[
hidelinks 
]{hyperref}
\usepackage{textcomp}
\def\BibTeX{{\rm B\kern-.05em{\sc i\kern-.025em b}\kern-.08em
    T\kern-.1667em\lower.7ex\hbox{E}\kern-.125emX}}

\usepackage[backend=biber, style=ieee, sorting=none, natbib=true, mincitenames=1, maxcitenames=2]{biblatex}
\bibliography{bib/references.bib}
\AtEveryBibitem{%
  \ifboolexpr{test {\ifentrytype{article}} or test{\ifentrytype{inproceedings}}}
    {\clearfield{url}%
     \clearfield{urlyear}}
    {}%
}

\usepackage{acronym}
\acrodef{ESDF}{Euclidean Signed Distance Field}
\acrodef{TSDF}{Truncated Signed Distance Field}
\acrodef{SDF}{Signed Distance Field}
\acrodef{ROS}{Robot Operating System}

\usepackage{soul}
\setstcolor{blue}

\usepackage{booktabs}
\usepackage{array}
\newcolumntype{M}[1]{>{\centering\arraybackslash}m{#1}}
\usepackage{siunitx} 

\begin{document}
\history{Date of current version July 6, 2023.}
\doi{}

\title{Volumetric Occupancy Detection: A Comparative Analysis of Mapping Algorithms}
\author{Manuel Gomes\authorrefmark{1,2},
Miguel Oliveira\authorrefmark{1,2}, and Vítor Santos\authorrefmark{1,2}
\IEEEmembership{Member, IEEE}}

\address[1]{Intelligent System Associate Laboratory (LASI),
Institute of Electronics and Informatics Engineering of Aveiro (IEETA), 
University of Aveiro, 3810-193 Aveiro, Portugal}
\address[2]{Department of Mechanical Engineering, University of Aveiro, 3810-193 Aveiro, Portugal}

\tfootnote{This work was supported by the Project Augmented Humanity [POCI-01-0247-FEDER-046103], financed by Portugal 2020, under the Competitiveness and Internationalization Operational Program, the Lisbon Regional Operational Program, and by the European Regional Development Fund.
\\
This work has been submitted to the IEEE for possible publication. Copyright may be transferred without notice, after which this version may no longer be accessible.
}

\markboth
{Manuel Gomes \headeretal: Volumetric Occupancy Detection: A Comparative Analysis of Mapping Algorithms}
{Manuel Gomes \headeretal: Volumetric Occupancy Detection: A Comparative Analysis of Mapping Algorithms}

\corresp{Corresponding author: Manuel Gomes (e-mail: manuelgomes@ua.pt).}

\begin{abstract}
Despite the growing interest in innovative functionalities for collaborative robotics, volumetric detection remains indispensable for ensuring basic security. 
However, there is a lack of widely used volumetric detection frameworks specifically tailored to this domain, and existing evaluation metrics primarily focus on time and memory efficiency. 
To bridge this gap, the authors present a detailed comparison using a simulation environment, ground truth extraction, and automated evaluation metrics calculation. 
This enables the evaluation of state-of-the-art volumetric mapping algorithms, including OctoMap, SkiMap, and Voxblox, providing valuable insights and comparisons through the impact of qualitative and quantitative analyses. 
The study not only compares different frameworks but also explores various parameters within each framework, offering additional insights into their performance.
\end{abstract}


\titlepgskip=-21pt

\maketitle

\section{Introduction}\label{sec:introduction}
The field of human-robot collaboration is experiencing rapid advancements, leading to the introduction of new functionalities and interfaces. 
One prominent area of development involves the use of visual interfaces for gesture recognition. 
Researchers have explored various approaches, including those utilizing RGB \cite{Baptista2023, Brock2020}, Depth \cite{Li2019} or Stereo cameras \cite{Jiang2019} to capture and interpret human gestures.
Another avenue of exploration in human-robot collaboration is the utilization of audio-based functionalities. 
Some researchers have focused on transforming speech into text \cite{Nassif2019} or detecting emotions from speech \cite{Abdelhamid2022}. 
Combining both visual and audio cues, certain authors have investigated the fusion of images and audio for speech \cite{Afouras2022} and emotion recognition \cite{Heredia2022}.
In addition to visual and audio interfaces, electromyography interfaces have emerged as a promising approach for gesture recognition in human-robot collaboration. 
These interfaces leverage the measurement of muscle activity signals to detect and interpret human gestures \cite{Cote-Allard2019, Qi2019, Simao2019}.
Furthermore, advanced functionalities in human-robot collaboration encompass the handover of objects between humans and machines. 
Researchers have proposed innovative strategies and techniques to facilitate smooth and efficient object transfer \cite{Castro2023, Peringal2022, Castro2021}.

While these functionalities and interfaces are intriguing from a research perspective, it is important to note that they are not indispensable for collaboration to take place. 
However, an essential functionality that underpins collaboration scenarios is volumetric detection. 
Volumetric detection provides a crucial security feature by enabling real-time detection of foreign objects within a workspace, including their current pose. 
While 1-D or 2-D barriers can detect the entrance of foreign objects, volumetric detection creates a proliferate environment for collaboration, enhancing safety and efficiency in collaborative settings.

Currently, there is a lack of widely adopted volumetric detection frameworks specifically designed for human-robot collaboration. 
Mapping frameworks serve as the foundation for most volumetric detection systems since volumetric detection is required for mapping purposes. 
However, the evaluation of mapping frameworks often lacks comprehensive and insightful metrics, with performance assessments primarily focused on time and memory efficiency indicators \cite{Hornung2013,DeGregorio2017,Oleynikova2017}.

This paper presents aims to address the aforementioned gaps in research. 
The objectives of this study are to:
\begin{itemize}
    \item Create a simulation environment capable of testing the capabilities of various volumetric detection frameworks;
    \item Establish a methodology to extract ground truth data from the simulation environment;
    \item Apply evaluation metrics and automate their calculation to enable a comprehensive evaluation;
    \item Provide insights into the performance of well-known mapping algorithms, namely OctoMap, SkiMap, and Voxblox, and identify optimal parameter settings for these frameworks.
\end{itemize}

The article is organized in five sections to comprehensively address these objectives. 
Introduction (\autoref{sec:introduction}) presents the purpose of this research and highlights the contributions of the authors. 
Related Work (\autoref{sec:related_work}) provides a detailed description of OctoMap, SkiMap, and Voxblox, including their potential advantages and disadvantages. 
Proposed Approach (\autoref{sec:proposed_approach}) outlines the testing environment, ground truth extraction methodology, and evaluation metrics in a comprehensive manner. 
Results (\autoref{sec:results}) offers an in-depth description of the experimental methodology employed, along with extensive qualitative and quantitative results. 
Finally, Conclusions (\autoref{sec:conclusions}) summarizes the approach and its advantages, while also delineating avenues for future research and development.
\section{Related Work}\label{sec:related_work}
This section provides an overview of three open-source mapping frameworks, namely OctoMap, SkiMap, and Voxblox. 
These mapping frameworks have played a crucial role in advancing the field of 3D mapping, empowering robots and autonomous systems to navigate and operate with efficiency and accuracy in complex, spatial environments.

\begin{figure}[ht!]
    \centering
    \includegraphics[width=0.8\columnwidth]{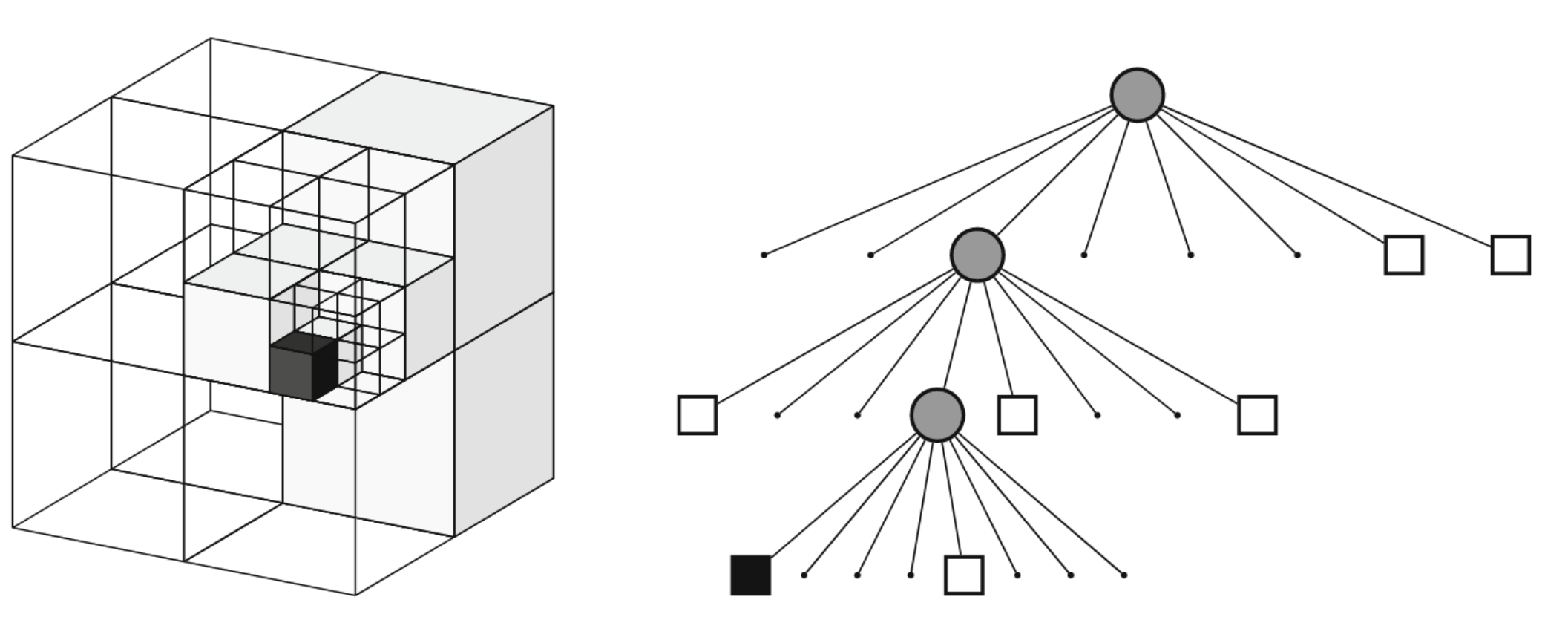}
    \caption{An octree that stores free cells, shaded in white, and occupied cells, colored black. The left side displays the volumetric model, while the right side exhibits the corresponding tree representation \cite{Hornung2013}.}
    \label{fig:octree}
\end{figure}

OctoMap is a mapping framework that utilizes a tree-based representation known as octrees to provide high flexibility in terms of mapped area and resolution \cite{Hornung2013}.
Octrees are hierarchical data structures for spatial subdivision in three dimensions \cite{Wilhelms1992}. 
Each octree consists of nodes that represent cubic volumes of space. 
In OctoMap, this tree is dynamically updated in real-time based on sensor measurements or observations. 
The tree begins with a single root node representing the entire space. 
Then, the node is recursively divided into eight more nodes until it reaches a predefined minimum voxel size, thereby discretizing the three-dimensional space into a hierarchical grid.
An example of an octree representation is depicted in \autoref{fig:octree}.

\begin{figure}[hb!]
    \centering
    \includegraphics[width=0.8\columnwidth]{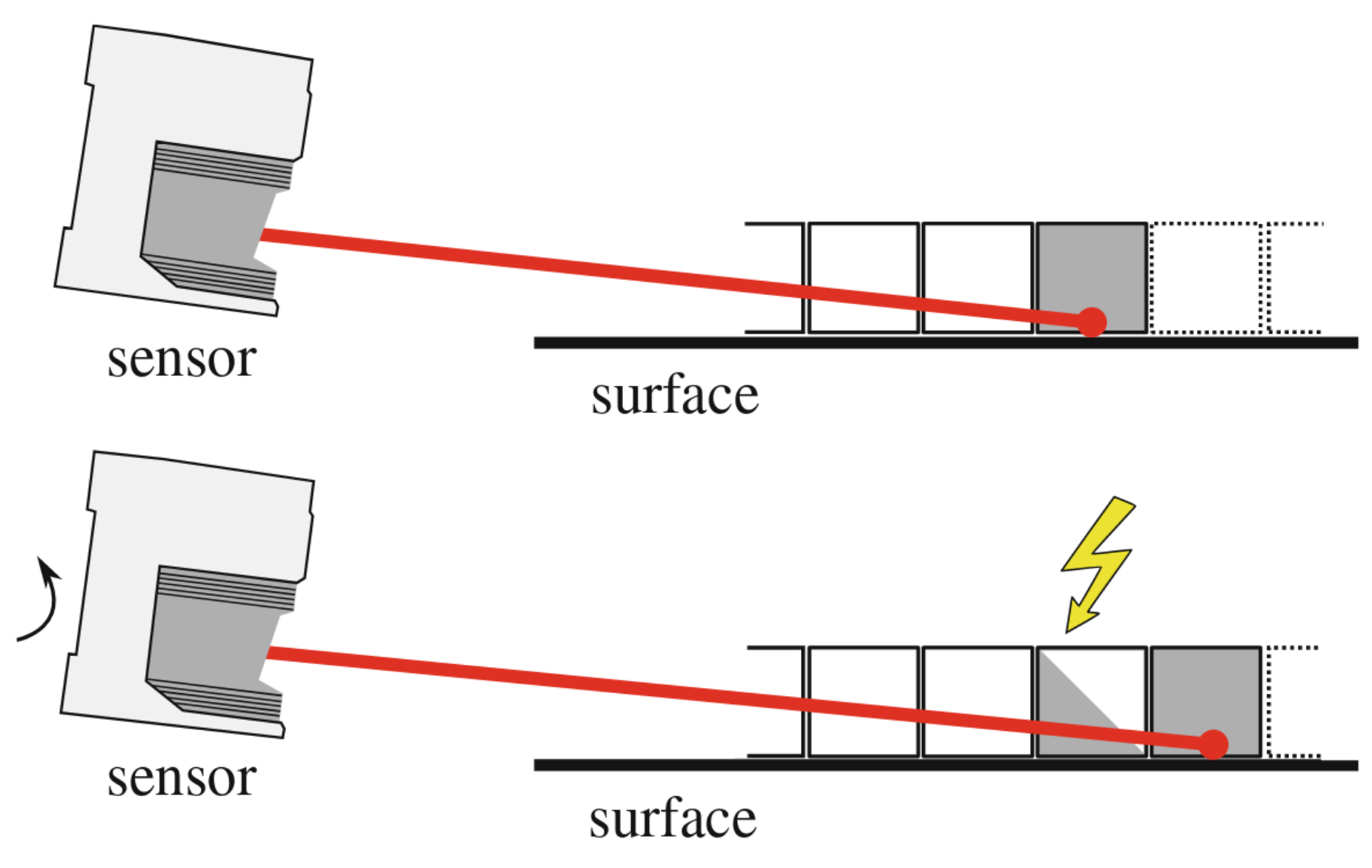}
    \caption{A laser scanner scans a flat surface at a shallow angle by rotating. In the first scan (top), a cell is detected as occupied, but in the subsequent scan (bottom) after the rotation of the sensor, the same cell is updated as free. Occupied cells are depicted as gray boxes, while free cells are represented in white \cite{Hornung2013}.}
    \label{fig:raycasting}
\end{figure}

The basic form of an octree can be used to model a Boolean property, where each node can have two states: occupied volume or free volume. 
In the case of volumetric detection, a node is initialized if a certain volume is occupied. 
However, this approach creates ambiguity by failing to differentiate between unknown and free voxels. 
To address this issue, OctoMap explicitly represents free volumes in the tree by employing raycasting. 
This technique assumes the volumes between the sensor and the measured endpoint as free space. 
A visual depiction of raycasting is presented in \autoref{fig:raycasting}.
By adopting this representation, the framework achieves a compact environment representation by pruning all children of the same node if they have the same state.

In scenarios involving sensor noise and dynamic environments, the Boolean tree representation is inadequate. 
Therefore, OctoMap incorporates a probabilistic model to represent occupancy. 
In this model, each node is associated with a probability value indicating the likelihood of a voxel being occupied. 
A threshold is often applied to classify a voxel as occupied or free based on its probability value. 
OctoMap adopts a holistic approach by considering the complete unfolding of a scene, rather than relying solely on time snapshots. 
Consequently, it incorporates multiple observations to determine the present state of a voxel. 
For instance, when a voxel has been observed as unoccupied in $k$ instances, a minimum of $k$ observations of occupancy is required to transition it from an unoccupied to an occupied state.
This characteristic strengthens the framework's resilience to sensor noise, facilitates temporal scene analysis, and enables sensor fusion. 
Nevertheless, in dynamic environments, this behavior becomes undesirable as it hinders the algorithm's responsiveness.
To address this issue, OctoMap proposes a clamping update policy that establishes upper and lower bounds on the number of observations needed to change the state.
In summary, OctoMap provides a powerful framework for 3D occupancy mapping, offering efficient storage, real-time updates, and probabilistic representation. Its octree-based structure enables accurate modeling of the environment, making it a valuable tool for robots and autonomous systems operating in three-dimensional spaces.

SkiMap is a mapping framework, akin to OctoMap, that employs a tree of SkipLists as its underlying data structure \cite{DeGregorio2017}. 
SkipLists are layered probabilistic data structures \cite{Pugh1990}. 
The first layer comprises an ordered linked list, and each subsequent layer is a subset of the layer below it, encompassing a fraction of the elements. 
This design reduces the computational complexity associated with random access.

\begin{figure}[hb!]
    \centering
    \includegraphics[width=0.7\columnwidth]{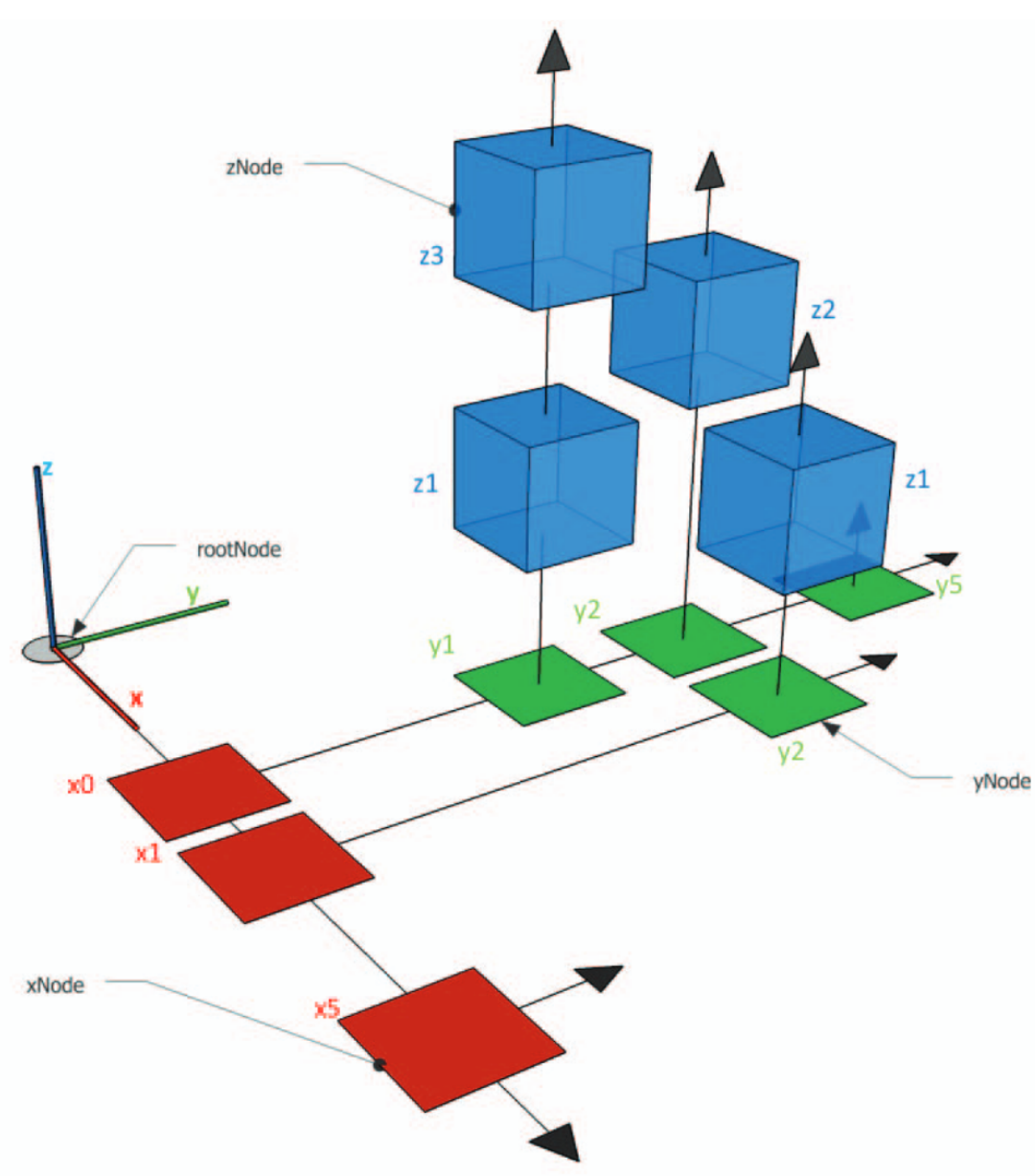}
    \caption{Voxel grouping process into a Tree of SkipLists. Each voxel, depicted as a blue box, is connected to the rootNode through a yNode, represented by a green tile, and the yNode is further linked to an xNode, depicted as a red tile \cite{DeGregorio2017}.}
    \label{fig:skiplists}
\end{figure}

In SkiMap, the tree of SkipLists consists of multiple levels, where the first SkipList tracks quantized x coordinates, representing depth level 1, and the individual elements in this SkipList are referred to as \textit{xNodes}. 
Each \textit{xNode} itself corresponds to a SkipList that tracks quantized y coordinates, representing depth level 2, with its elements named \textit{yNodes}. 
Finally, each \textit{yNode} corresponds to a SkipList of \textit{zNodes}, representing the actual voxels and serving as containers for user data.
As a result of this data structure, voxel coordinates can be obtained by traversing predecessors, obviating the need to store coordinates within the containers alongside user data. 
This tree of SkipLists is better visualized in \autoref{fig:skiplists}.

To classify voxels as occupied or free, Each voxel within the tree of SkipLists possesses a dynamically defined weight. 
When a detection occurs within a voxel, its weight increases, while it decreases when no detection is present. 
If the weight surpasses a predefined threshold known as the minimum voxel weight, the voxel is classified as occupied.
The authors of SkiMap argue that their methodology leads to reduced runtime and memory usage compared to OctoMap.

Voxblox is a mapping framework similar to the previously discussed OctoMap and SkiMap \cite{Oleynikova2017}. 
Originally developed for Micro Aerial Vehicles operating in unstructured and unexplored environments, Voxblox provides a fast and flexible local mapping solution.

This framework is based on the creation of \acp{ESDF} derived from \acp{TSDF}. 
Both \acp{ESDF} and \acp{TSDF} are data structures built upon \acp{SDF}, which are voxel grids that divide the 3D space. 
Each voxel is assigned a projective distance label representing the distance between the voxel and the nearest detected surface. 
Positive and negative values indicate whether the voxel is located outside or inside the surface, respectively.

\begin{figure}[hb!]
    \centering
    \includegraphics[width=0.6\columnwidth]{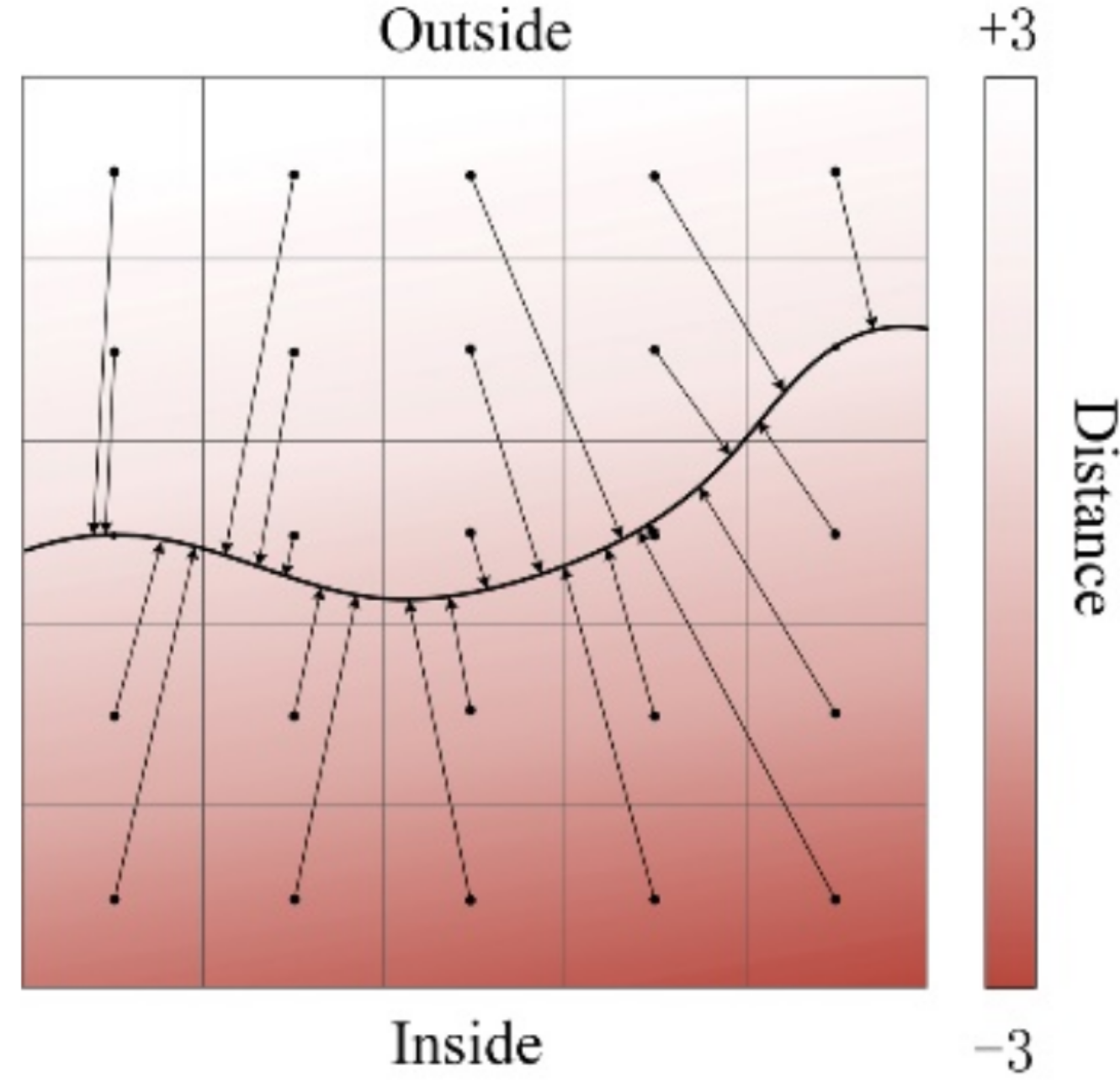}
    \caption{2D cross-section of a general \acf{TSDF} computed on a grid of regular voxels, with each arrow representing the distance vector of a voxel \cite{He2020}.}
    \label{fig:tsdf}
\end{figure}

In \acp{TSDF}, a truncation distance is defined to limit the storage of distances, optimizing memory usage and processing speed by focusing on the relevant parts of the environment. 
A \ac{TSDF} representation is presented in \autoref{fig:tsdf}.
In contrast, \acp{ESDF} employ the Euclidean distance between the voxel and the nearest obstacle instead of the projective distance. 
Consequently, \acp{ESDF} are particularly useful for navigation purposes.

Voxblox calculates \acp{TSDF} by employing raycasting and averaging the newly measured projective distances into the existing voxels. 
For each detection, a weighting function is applied, which decreases quadratically with the distance from the sensor to the surface and linearly with the distance from the voxel to the surface. 
Subsequently, Voxblox derives \acp{ESDF} from \acp{TSDF} using a method based on occupancy maps \cite{Lau2010}. 
This method utilizes wavefronts, propagating waves that update distances from the start voxel to its neighbors. 
Updated voxels are added to the wavefront queue for further propagation.
Voxblox employs two wavefronts: a raise wavefront and a lower wavefront. 
The raise wavefront is triggered when the new distance value of a voxel from the \ac{TSDF} exceeds the previously stored value in the corresponding \ac{ESDF} voxel. 
Invalidation of the voxel and its children follows, as they are added to the raise queue. 
The wavefront continues propagation until no voxels with invalidated parents remain. 
Conversely, the lower wavefront initiates when a new occupied voxel enters the map or when a previously observed voxel decreases its value. 
Based on the current voxel, its neighbors, and their respective distances, the distances of neighboring voxels are updated. 
The wavefront propagation terminates when there are no remaining voxels whose distances could decrease based on their neighbors.

The authors of Voxblox argue that constructing a \ac{TSDF} is faster than building an OctoMap, and constructing an \ac{ESDF} from a \ac{TSDF} is faster than building it from an occupancy map.

\section{Proposed Approach}\label{sec:proposed_approach}
The frameworks described in the previous section are usually assessed using only basic metrics, such as time of integration of voxel grid or memory usage \cite{DeGregorio2017,Oleynikova2017}.
While these metrics serve as valuable indicators of efficiency, they prove insufficient for volumetric representation. 
The existing body of research lacks a comprehensive measurement approach that adequately assesses the alignment between the outputs of frameworks and the actual objects being detected. 
Consequently, the authors of this study have undertaken the development of a tool specifically designed to carry out such evaluation. 
The subsequent sections will provide a detailed description of this tool.

\subsection{Testing Environment}\label{subsec:testing_environment}
The development of a simulation-based testing environment by the authors was a strategic choice, primarily driven by the objective of acquiring ground truth values, which cannot be feasibly obtained in a real environment. 
Although this decision effectively eliminated the challenges associated with hardware-related issues encountered in testing with physical sensors, its primary motivation remains the acquisition of accurate reference data.

Gazebo\footnote{\url{https://gazebosim.org}}, an open-source robotics simulator, is used to construct their simulation environment. 
It was selected due to its attributes, including a highly accurate physics engine that faithfully replicates object behavior. 
Furthermore, Gazebo has the capability to simulate multiple sensors concurrently, allowing for the introduction of sensor noise. 
A noteworthy feature of Gazebo is its capacity to animate static models, referred to as actors, which plays a vital role in creating realistic environments.
Moreover, Gazebo's integration with the \ac{ROS} is particularly significant. 
\ac{ROS}, an open-source framework, offers a comprehensive set of software libraries and tools that are instrumental in the development, control, and integration of robotic systems.
This integration is important because all the algorithms employed in this study are \ac{ROS}-integrated. 
This seamless integration significantly enhances the ease-of-use and compatibility of the simulation environment with the algorithms, ensuring an efficient workflow.

\begin{figure}[ht!]
    \centering
    \includegraphics[width=\columnwidth]{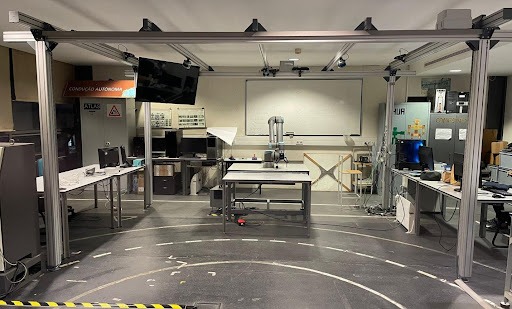}
    \caption{LARCC, an advanced collaborative cell embedded with an extensive array of sensory mechanisms.}
    \label{fig:larcc-real}
\end{figure}

The simulated environment was based on LARCC, a collaborative cell equipped with three LiDARs, one RGB-D camera, and three RGB cameras \cite{Rato2022}. 
This real-life setup can be observed in \autoref{fig:larcc-real}. 
The environment includes four sensors suitable for volumetric detection: the three LiDARs and the RGB-D camera.
The dimensions of the cell are 4 meters in length, 2.8 meters in width, and 2.29 meters in height.

\begin{figure}[hb!]
    \centering
    \includegraphics[width=\columnwidth]{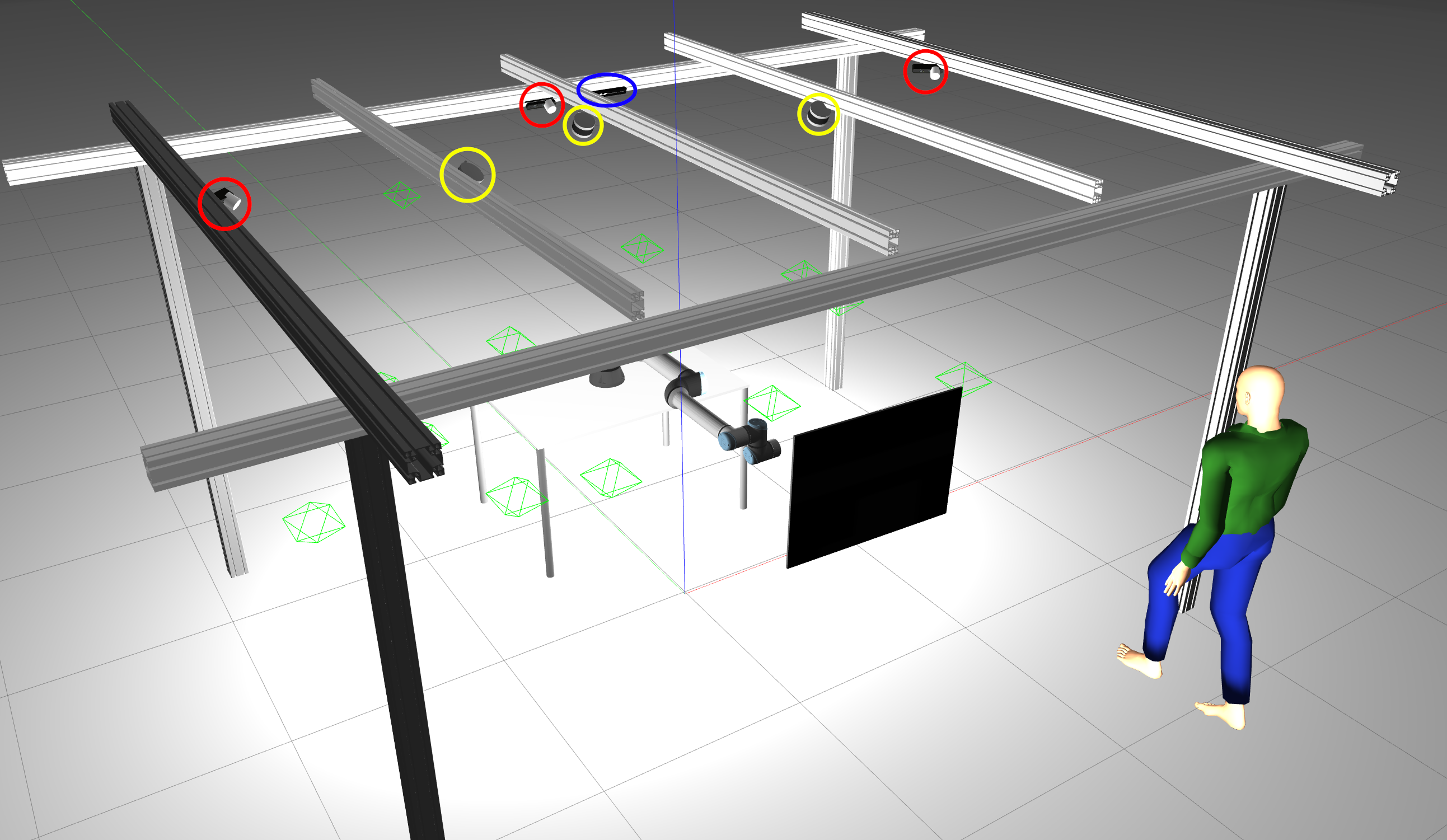}
    \caption{Testing environment, representing LARCC. The cell contains 3 RGB cameras, 3 LiDARs and 1 RGB-D, signaled by red, yellow and blue ellipses, respectively.
    The actor is presented as a human with a green shirt.}
    \label{fig:larcc}
\end{figure}

The purpose of this environment is to test the ability of occupancy algorithms to promptly detect a person entering the cell. 
To simulate this scenario accurately, a Gazebo actor representing a person was spawned in the middle of the cell. 
The actor moves in an elliptical pattern along the center of the cell, with a semi-major axis of 2 meters aligned with the length of the cell and a semi-minor axis of 1 meter aligned with the width. 
The actor moves in this pattern two times, totaling into an episode of around 96 seconds.
In the center of this elliptical movement, a rectangular plate measuring 0.8 meters in length, 0.6 meters in width, and 1 meter in height was placed. 
This plate serves to evaluate the algorithms' performance when the moving object is partially occluded.
The environment can be seen in detail in \autoref{fig:larcc}.

\subsection{Data Filtering and Retrieval}\label{subsec:filtering}
The testing environment produces a point cloud for each range-finder sensor. 
To conduct an evaluation, it is essential to obtain the ground truth data that correspond to the actor. 

The fusion of point clouds from multiple sensors is a prerequisite for accurate evaluation. 
To achieve this, an approximate time filter is implemented\footnote{\url{https://wiki.ros.org/message_filters/ApproximateTime}}. 
This filter organizes the ROS messages into a queue for each sensor.
When all four sensors to have a point cloud message in the queue, the filter retrieves the latest message from each sensor and then clears the queue.

In order to enhance the efficiency of the testing framework, the authors implemented a downsampling technique on the point cloud acquired from the RGB-D camera. 
The point cloud exhibited high density, resulting in computationally intensive operations. 
To mitigate this issue, a mean pooling algorithm was employed to downsample the point cloud.

In this approach, the LinkStates message provided by Gazebo is utilized\footnote{\url{https://docs.ros.org/en/jade/api/gazebo_msgs/html/msg/LinkStates.html}}. 
The LinkStates message contains real-time pose information for each rigid body (link) present in the simulation. 
The actor consists of various links, such as "Head", "RightForeArm", "LowerBack", "LeftUpLeg", among others. 
To determine the connections between these links, an automated analysis of the original mesh file (.dae) of the actor is performed.

A three-dimensional volume for each connection (joint) that encapsulates the entire link within it was created. 
For instance, a cylinder with a radius of 0.1 meters is defined between the shoulder and the elbow. 
A more complex example involves a rectangular prism with a width of 0.2 meters and a length of 0.3 meters between the Spine and the LowerBack. 
One of the prism's axes is aligned with the line connecting the LeftHip and the RightHip.

With these volumes established, point-in-polyhedron algorithms were implemented to extract the actor points from the original point clouds. 
For a cylinder with a radius of $r$, the algorithm verifies whether a point, represented by the coordinate vector $\vec{q}$, lies between the top and bottom faces of the cylinder. 
The centers of these faces are represented by the coordinate vectors $\vec{p}_1$ and $\vec{p}_2$, respectively. 
Additionally, the algorithm checks if the point lies within the curved surface of the cylinder. 
This is done by verifying if the distance between the point and the cylinder's center axis is smaller than the radius $r$.
These operations are illustrated by \autoref{eq:cyl1}, \autoref{eq:cyl2}, and \autoref{eq:cyl3}.

\begin{equation}
    (\vec{q}-\vec{p}_1)\cdot(\vec{p}_2-\vec{p}_1) \geq 0
    \label{eq:cyl1}
\end{equation}

\begin{equation}
    (\vec{q}-\vec{p}_2)\cdot(\vec{p}_2-\vec{p}_1) \leq 0
    \label{eq:cyl2}
\end{equation}

\begin{equation}
    \frac{|(\vec{q}-\vec{p}_1)\times(\vec{p}_2-\vec{p}_1)|}{|\vec{p}_2-\vec{p}_1|} \geq r
    \label{eq:cyl3}
\end{equation}

For a rectangular prism with dimensions of length $l$, width $w$, and height $h$, the normal vectors for all its faces are computed. 
Specifically, $\vec{n}_l$, $\vec{n}_w$, and $\vec{n}_h$ represent the normal vectors along the length, width, and height of the prism, respectively. 
To determine if a point $q$ lies within the prism, \autoref{eq:rect} is utilized.
This equation calculates the coordinate vector of the point with respect to the coordinate vector of the prism's center, denoted as $\vec{c}$. 
Subsequently, the dot product of this vector with each of the normal vectors is computed. 
This calculation enables an assessment of how far the point is from the center in a particular direction. 
If the result is greater than half of the corresponding dimension of the prism, the point is considered outside the prism.

\begin{equation}
    |(\vec{q} - \vec{c})\cdot\vec{n}_{\alpha}| \leq \frac{\alpha}{2},\quad \alpha \in \{l,w,h\}
    \label{eq:rect}
\end{equation}

\begin{figure}[hb!]
    \centering
    \includegraphics[width=\columnwidth]{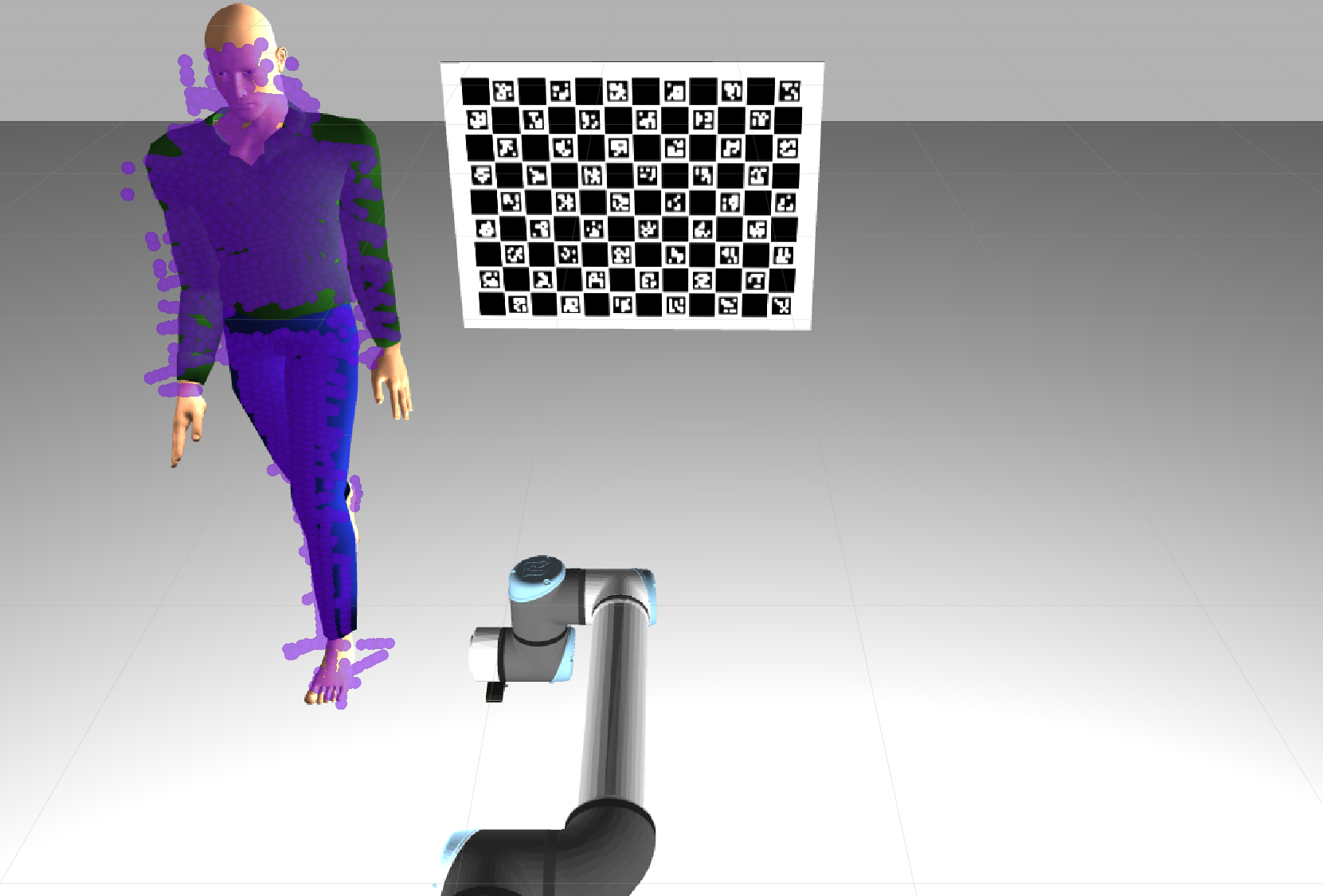}
    \caption{Projection of actor points, in purple, to the image of the center camera of LARCC.}
    \label{fig:clustering}
\end{figure}

From this method, the actor points can be extracted, as seen in \autoref{fig:clustering}.
However, a challenge arises due to the mismatched units of measurement between this points and the voxels retrieved from the mapping framework. 
To address this issue, a voxel grid was constructed with matching resolution and pose as the voxel grid defined by the framework. 
In this new grid, occupied voxels represent the locations of the actor points, therefore being the ground truth data.

\subsection{Evaluation Metrics}\label{subsec:metrics}
The assessment of the effectiveness of the framework will be predicated upon several performance metrics, namely Precision ($P$), Recall ($R$), and F1 Score ($F_1$).
The inclusion of $F_2$ and $F_3$ scores are also presented, as these scores place a higher emphasis on recall in relation to precision. 
This emphasis proves valuable, particularly in contexts involving security scenarios, such as volumetric detection in a factory setting.
These metrics are defined by \autoref{eq:precision}, \autoref{eq:recall}, and \autoref{eq:fb-score}, in which $TP$, $FP$, and $FN$ are true positives, false positives, and false negatives, respectively. 
In \autoref{eq:fb-score}, the $\beta$ is 1 in $F_1$ score, 2 in $F_2$ score and 3 in $F_3$ score.

\begin{equation}
    P = \frac{TP}{TP+FP}
    \label{eq:precision}
\end{equation}

\begin{equation}
    R = \frac{TP}{TP+FN}
    \label{eq:recall}
\end{equation}

\begin{equation}
    F_{\beta} = (1+\beta^2)\cdot\frac{P \cdot R}{(\beta^2 \cdot P) + R}
    \label{eq:fb-score}
\end{equation}

To establish $TP$, $FP$ and $FN$ values within the context of this work, it is necessary to define relevant and retrieved elements. 
Relevant elements refer to the ground truth voxels that have been previously extracted, as described in \autoref{subsec:filtering}. 
Retrieved elements, on the other hand, correspond to the output of the frameworks in the form of voxels. 

Ensuring the retrieved elements are extracted after relevant elements is crucial. 
To accomplish this temporal alignment, an approximate time filter, resembling the one described in \autoref{subsec:filtering}, is implemented.
This filter organizes the output messages from the mapping framework into a queue, arranging them in ascending order based on their timestamps. 
It identifies the most recent timestamp among the fused point cloud messages. 
Consequently, it retrieves the earliest message from the mapping framework queue that is received after point cloud messages latest timestamp.

\begin{figure}[hb!]
    \centering
    \includegraphics[width=\columnwidth]{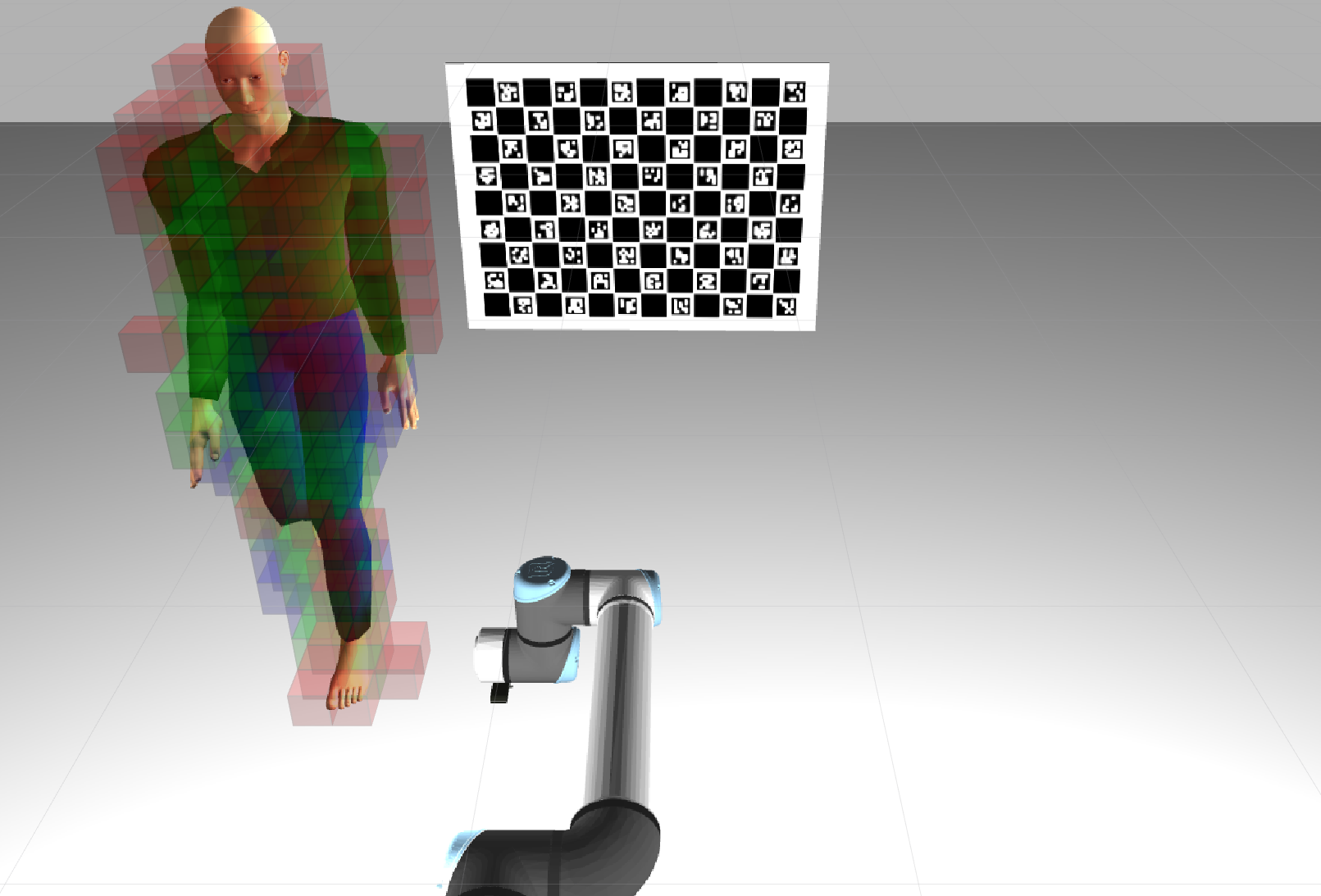}
    \caption{Projection of true positive, false positive, and false negative voxels to the image of the center camera of LARCC. The voxels are represented in green, blue, and red, respectively.}
    \label{fig:metrics}
\end{figure}

With both the relevant and retrieved elements established, $TP$ can be identified as voxels that are classified as occupied by both the framework's and the ground truth's voxel grid. 
$FP$ present a greater challenge, as the framework's voxel grid encompasses the entire scene rather than solely focusing on the actor. 
To address this, the convex hull of the actor points is obtained, and any retrieved element within this convex hull that is not a $TP$ is classified as $FP$. 
Conversely, $FN$ are voxels that are identified as occupied by the ground truth but classified as unoccupied by the frameworks. 
By utilizing these definitions, the aforementioned metrics can be calculated to quantitatively assess the performance of the framework.
These definitions can be visualized in \autoref{fig:metrics}.

\section{Results}\label{sec:results}
This section presents the experimental evaluation and results of the OctoMap, SkiMap, and Voxblox frameworks. 
It encompasses both qualitative and quantitative analyses to comprehensively assess the capabilities and limitations of these frameworks in capturing and representing the dynamic environment. 
The experimental infrastructure is described, followed by the qualitative results of the default parameter settings for each framework. 
Finally, the quantitative results, including precision, recall, and F scores, are presented and analyzed, offering valuable insights into the performance of OctoMap, SkiMap, and Voxblox.

\subsection{Experimental Methodology}\label{subsec:experimental-methodology}
In order to standardize the experiments, a ROS tool named bag files was utilized\footnote{\url{https://wiki.ros.org/Bags}}. 
These files are capable of capturing all the ROS messages, such as sensor pointclouds, within a specific time frame. 
Subsequently, they allow for the replaying of the recorded messages at the original recording rate. 
By recording the environment described in \autoref{subsec:testing_environment}, it ensures consistency and minimizes variability between experiments.
Bag files offer the functionality to adjust the data reproduction speed, enabling both deceleration and acceleration. 
This feature proves particularly useful in this study, as slowing down the bag file playback allows the mapping framework sufficient time to integrate the map. 
This ensures that slower frameworks are not disadvantaged. 
It is important to note that the objective of this study is not primarily focused on measuring efficiency.

The experiments conducted not only involve comparing the default versions of the frameworks with each other but also exploring different parameters within each framework. 
Some parameters exhibit variations across frameworks. 
These parameters are: 
\begin{itemize}
    \item Minimum Voxel Length --- This parameter represents the length of the smallest voxel into which the algorithm can divide the environment.
    \item Hit Probability --- It denotes the probability of the algorithm increasing the occupied probability of a voxel that has a detection.
    \item Miss Probability --- Probability of the algorithm reducing the occupied probability assigned to a voxel lacking a detection. 
    \item Minimum Voxel Weight --- It indicates the minimum weight required for a voxel to be classified as occupied.
    \item Truncation Distance --- This value represents the maximum distance considered for \ac{TSDF} calculations in Voxblox.
    \item Constant Weight --- When this boolean parameter is active, calculated weights remain constant irrespective of the distance between the point and the sensor.
\end{itemize}
The group of parameters and their default values can be summarized in \autoref{tab:params}.


\begin{table}[!ht]
\caption{Test parameters and their default values for each of the tested frameworks: OctoMap, SkiMap and Voxblox. N/A means not applicable.}
\label{tab:params}
\centering
\begin{tabular}{lccc}
\toprule
                         & OctoMap & SkiMap & Voxblox\\
\midrule
Minimum Voxel Length (m) & 0.1     & 0.1    & 0.1   \\ 
Hit Probability          & 1       & N/A    & N/A   \\
Miss Probability         & 0.4     & N/A    & N/A   \\
Minimum Voxel Weight     & N/A     & 1      & N/A   \\
Truncation Distance (m)  & N/A     & N/A    & 0.1   \\
Constant Weight          & N/A     & N/A    & False \\ \bottomrule
\end{tabular}
\end{table}

\subsection{Qualitative Results}\label{subsec:qualitative-results}
In this subsection, the authors aim to provide qualitative results of the frameworks utilizing their default parameters. 
All figures are consistently captured at the same moment in time and scene position, specifically when the actor passes for the second time between the middle plate and the right leg of the gantry. 

\begin{figure}[!hb]
    \centering
    \includegraphics[width=\columnwidth]{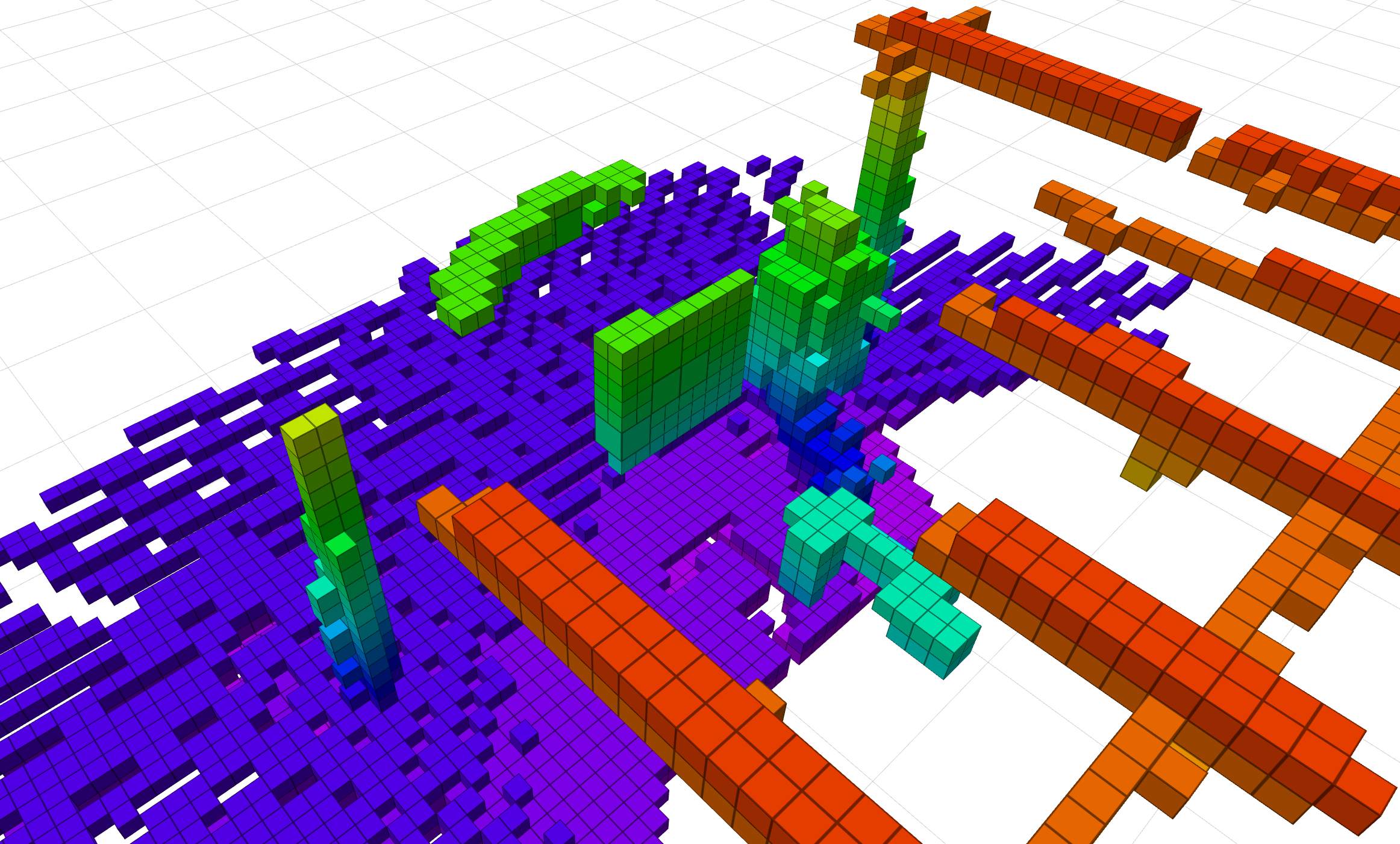}
    \caption{OctoMap occupied voxels representing the actor passing through between the middle plate and the right leg of the gantry.}
    \label{fig:octomap-qualitative}
\end{figure}

Commencing with OctoMap, as depicted in \autoref{fig:octomap-qualitative}, the actor's representation is clearly discernible, and the entire scene is well-defined. 
However, a limitation arises wherein certain voxels that previously indicated occupancy encounter difficulty in accurately depicting emptiness, particularly those corresponding to the upper side of the actor when it traverses behind the plate. 
This phenomenon arises from the properties of raycasting, as the rays that intersect with these voxels do not encounter any objects behind them from the perspective of the sensors. 
Consequently, the affected voxels cannot be effectively cleared, leading to a disparity between their actual empty state and their visual representation.

\begin{figure}[!ht]
    \centering
    \includegraphics[width=\columnwidth]{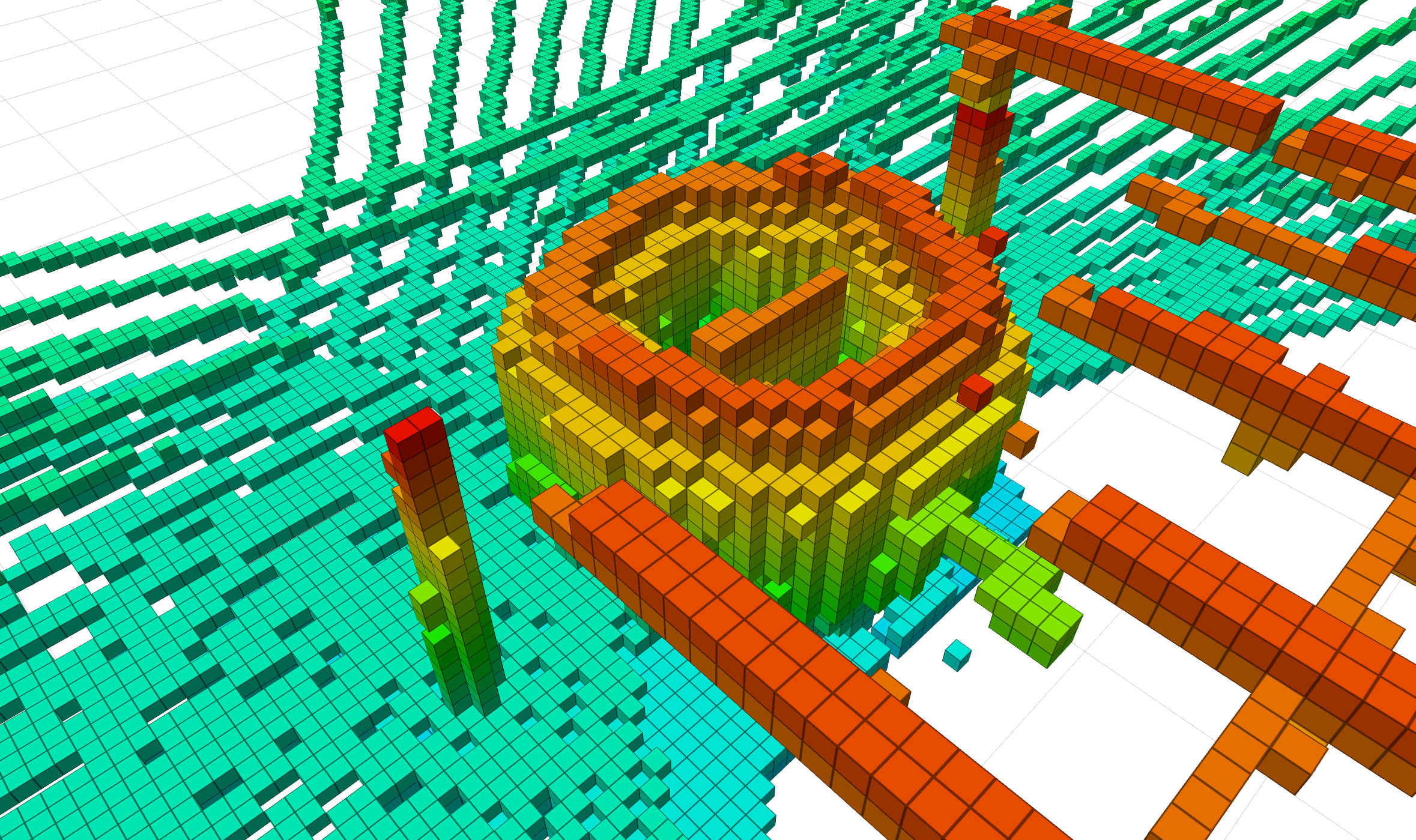}
    \caption{SkiMap occupied voxels representing the actor passing through between the middle plate and the right leg of the gantry.}
    \label{fig:skimap-qualitative}
\end{figure}

\autoref{fig:skimap-qualitative} illustrates the qualitative results of SkiMap. 
Interestingly, the issue observed in OctoMap is magnified in this framework, as there appears to be minimal to no occupied voxel erasure. 
This occurrence can be attributed to the fact that SkiMap was not specifically designed with volumetric detection in mind and, consequently, is ill-equipped to handle such scenarios. 
The framework's limitations in effectively handling and updating voxel occupancy representations in dynamic environments become more apparent, further highlighting its lack of readiness for addressing these specific types of situations.

\begin{figure}[!hb]
    \centering
    \includegraphics[width=\columnwidth]{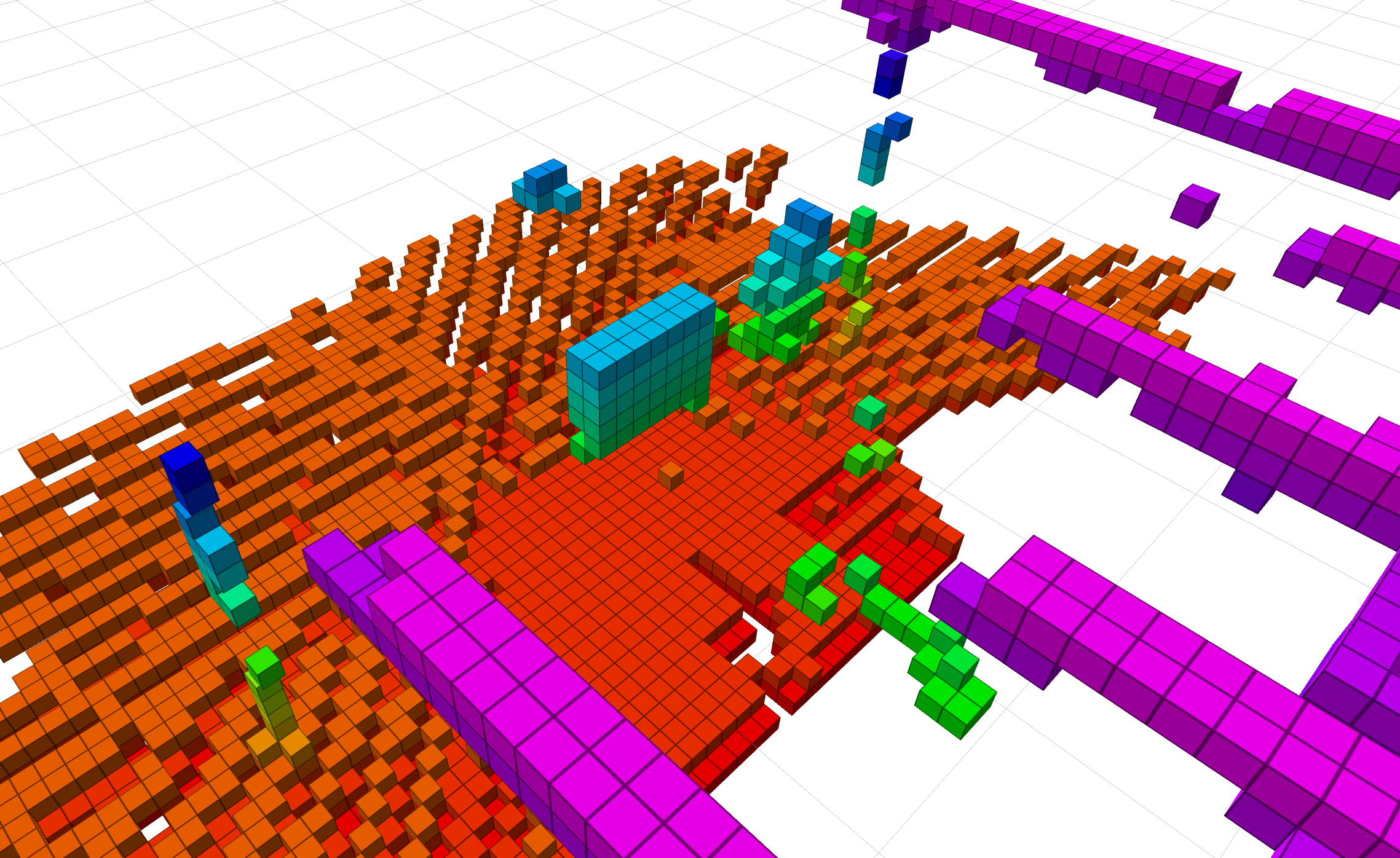}
    \caption{Voxblox occupied voxels representing the actor passing through between the middle plate and the right leg of the gantry.}
    \label{fig:voxblox-qualitative}
\end{figure}

\autoref{fig:voxblox-qualitative} presents the qualitative results of Voxblox. 
It is evident that this framework encounters significant challenges in detecting moving actors within the scene. 
This difficulty can potentially be attributed to the adoption of a high truncation distance, which leads to the framework observing a larger volume encompassing the actor. 
As a consequence, the larger volume primarily consists of empty space, resulting in a comparatively greater distance value assigned to the voxel and subsequently classifying it as empty. 
A more comprehensive explanation of this phenomenon can be found in \autoref{subsec:quantitative-results}, where detailed quantitative results are elaborated upon.

\subsection{Quantitative Results}\label{subsec:quantitative-results}
In this subsection, the authors present the quantitative results of precision, recall, and F scores obtained from evaluating different variations of OctoMap, SkiMap, and Voxblox. 
These metrics provide objective measures of the frameworks' performance in capturing and representing the dynamic environment. 
The analysis aims to provide deeper insights into the strengths and limitations of each framework, facilitating a comprehensive understanding of their respective capabilities.

Starting with OctoMap, the effect of the minimum voxel length can be seen in \autoref{tab:octomap-length}.
The precision consistently improves with higher lengths, possibly because false positives are generally situated in close proximity to true positives. 
By enlarging the voxels, this proximity is now encompassed within the true positive voxels, thereby enhancing precision. 
The change in recall, however, is less pronounced and reaches its peak at 0.1 meters. 
Generally, recall is smaller than precision due to occlusion in the testing environment, where the front half of the actor obstructs the back half, resulting in a considerable number of false negatives. 
Smaller lengths yield smaller recall as a smaller volume of the actor is covered by occupied voxels. 
Similarly, higher lengths also yield smaller recall because certain false negatives persist while the overall voxel count decreases, thereby elevating the false negatives to true positives ratio. 
Thus, the authors propose that an optimal recall point is reached in this environment, estimated to be around 0.1 meters. 
The authors further conclude that a length of 0.150 meters excels in applications with lower security requirements, while a length of 0.1 meter is ideal for scenarios prioritizing security.


\begin{table}[!ht]
\centering
\caption{Values of precision, recall and F scores for OctoMap when altering minimum voxel length.}
\label{tab:octomap-length}
\setlength{\tabcolsep}{3.5pt}
\begin{tabular}{S*{5}{c}}
\toprule
\multicolumn{1}{M{2cm}}{Minimum Voxel Length (m)} & Precision & Recall & F1 Score & F2 Score & F3 Score \\ \midrule
0.05       & 0.347     & 0.416  & 0.379    & 0.400    & 0.408    \\
0.075      & 0.519     & 0.468  & 0.492    & 0.477    & 0.473    \\
0.1        & 0.650     & \textbf{0.477}  & 0.550    & 0.504    & \textbf{0.490}    \\
0.125      & 0.750     & 0.468  & 0.577    & \textbf{0.506}    & 0.487    \\
0.150      & 0.833     & 0.460  & \textbf{0.593}    & \textbf{0.506}    & 0.482    \\
0.175      & 0.883     & 0.439  & 0.586    & 0.488    & 0.462    \\
0.2        & \textbf{0.940}     & 0.425  & 0.586    & 0.478    & 0.450   \\
\bottomrule
\end{tabular}%
\end{table}

The impact of OctoMap hit probability is demonstrated in \autoref{tab:octomap-hit} of this study. 
Notably, higher probabilities yield improved precision and recall measures, with precision values remaining constant for probabilities of 0.7 and 1. 
This trend can be attributed to the increased sensitivity of the framework to changes as the probability is increased, thereby reducing the occurrences of false negatives and false positives. 
Conversely, decreasing the probability results in an inaccurate representation of the environment, leading to an augmented number of false negatives and false positives. 
Based on these findings, the authors strongly advocate for a hit probability of 1 across all scenarios.


\begin{table}[!ht]
\centering
\caption{Values of precision, recall and F scores for OctoMap when altering hit probability.}
\label{tab:octomap-hit}
\setlength{\tabcolsep}{3.5pt}
\begin{tabular}{S*{5}{c}}
\toprule
\multicolumn{1}{M{2cm}}{Hit probability} & Precision & Recall & F1 Score & F2 Score & F3 Score \\ \midrule
0.5             & 0.414     & 0.011  & 0.021    & 0.014    & 0.012    \\
0.7             & \textbf{0.675}     & 0.218  & 0.330    & 0.252    & 0.234    \\
1               & 0.650     & \textbf{0.477}  & \textbf{0.550}    & \textbf{0.504}    & \textbf{0.490}  \\
\bottomrule
\end{tabular}%
\end{table}

The impact of OctoMap miss probability is illustrated in \autoref{tab:octomap-miss} of this investigation. 
Upon examination, it is observed that increasing the miss probability to 0.7 leads to a reduction in both precision and recall values. 
Conversely, when the probability fluctuates within the range of 0.2 to 0.4, the results remain relatively consistent. 
The decline in precision and recall can be attributed to a substantial decrease in the weight assigned to voxels, rendering them significantly less likely to be classified as occupied, even in the presence of detections. 
Consequently, this amplifies the number of false negatives and diminishes the count of true positives. 
Furthermore, this imbalance results in a larger number of false positives compared to true positives. 
Hence, for scenarios resembling the one described, the authors endorse employing a low miss percentage.


\begin{table}[!ht]
\centering
\caption{Values of precision, recall and F scores for OctoMap when altering miss probability.}
\label{tab:octomap-miss}
\setlength{\tabcolsep}{3.5pt}
\begin{tabular}{S*{5}{c}}
\toprule
\multicolumn{1}{M{2cm}}{Miss Probability} & Precision & Recall & F1 Score & F2 Score & F3 Score \\ \midrule
0.2              & \textbf{0.675}     & 0.459  & 0.547    & 0.491    & 0.475    \\
0.4              & 0.650     & \textbf{0.477} & \textbf{0.550}    & \textbf{0.504}    & \textbf{0.490}    \\
0.7              & 0.430     & 0.027  & 0.050    & 0.033    & 0.030   \\
\bottomrule
\end{tabular}
\end{table}

The impact of the minimum voxel length on SkiMap is demonstrated in \autoref{tab:skimap-length} of this study. 
The observed behavior is analogous to that observed in OctoMap, whereby precision increases as the voxel length enlarges. 
Additionally, the recall reaches its peak at approximately 0.075 meters in length. 
Based on these findings, the authors further infer that a voxel length of 0.150 meters is particularly well-suited for applications with lower security requirements. 
Alternatively, voxel lengths of 0.075 or 0.1 meters are recommended for scenarios that prioritize security considerations.


\begin{table}[!ht]
\centering
\caption{Values of precision, recall and F scores for SkiMap when altering minimum voxel length.}
\label{tab:skimap-length}
\setlength{\tabcolsep}{3.5pt}
\begin{tabular}{S*{5}{c}}
\toprule
\multicolumn{1}{M{2cm}}{Minimum Voxel Length (m)} & Precision & Recall & F1 Score & F2 Score & F3 Score \\ \midrule
0.05       & 0.310     & 0.566  & 0.400    & 0.485    & 0.523    \\
0.075      & 0.470     & \textbf{0.657}  & 0.548    & 0.608    & \textbf{0.632}    \\
0.1        & 0.606     & 0.629  & 0.617    & \textbf{0.624}    & 0.627    \\
0.125      & 0.707     & 0.581  & 0.638    & 0.603    & 0.592    \\
0.150      & 0.799     & 0.544  & \textbf{0.647}    & 0.581    & 0.562    \\
0.175      & 0.867     & 0.500  & 0.633    & 0.545    & 0.521    \\
0.2        & \textbf{0.922}     & 0.466  & 0.619    & 0.517    & 0.490   \\
\bottomrule
\end{tabular}
\end{table}

The impact of the minimum voxel weight on SkiMap is depicted in \autoref{tab:skimap-weight} of this investigation. 
It is observed that as the minimum weight increases, precision values also increase, while recall values decrease. 
This phenomenon can be attributed to the fact that higher minimum weights result in detections occurring only in voxels that exhibit a high level of certainty in terms of occupancy. 
Consequently, the number of false positives is reduced, but the number of false negatives is amplified. 
However, when evaluating the F scores, it is evident that a minimum weight of 1 yields the most optimal performance for this particular scenario.


\begin{table}[!ht]
\centering
\caption{Values of precision, recall and F scores for SkiMap when altering minimum voxel weight.}
\label{tab:skimap-weight}
\setlength{\tabcolsep}{3.5pt}
\begin{tabular}{S*{5}{c}}
\toprule
\multicolumn{1}{M{2cm}}{Minimum Voxel Weight} & Precision & Recall & F1 Score & F2 Score & F3 Score \\ \midrule
1 & 0.606     & \textbf{0.629}  & \textbf{0.617}    & \textbf{0.624}    & \textbf{0.627}    \\
5 & 0.613     & 0.582  & 0.597    & 0.588    & 0.585    \\
10 & \textbf{0.624}     & 0.542  & 0.580    & 0.557    & 0.549   \\
\bottomrule
\end{tabular}
\end{table}

The impact of the minimum voxel length on Voxblox is presented in \autoref{tab:voxblox-length}, highlighting similarities and notable distinctions from the findings of OctoMap and SkiMap. 
While some similarities in behavior are observed, a striking difference is identified. 
Specifically, for voxel lengths smaller than 0.150 meters, a significant decrease in recall is observed. 
This phenomenon can be attributed to the relationship between the truncation distance and the minimum voxel length, which will also be discussed in detail in the discussion of \autoref{tab:voxblox-truncation-distance}. 
The authors conducted an analysis and found that achieving a ratio of voxel size per truncation distance of at least 1.5 leads to an increase in recall, despite Voxblox's recommended ratio of 0.5. 
The authors speculate that this improvement occurs because as the ratio increases, the framework incorporates points that are farther away from the voxel, consequently yielding a larger average distance value for that voxel. 
This effect may generate distance values that fall below the occupied threshold, resulting in false negatives.


\begin{table}[!ht]
\centering
\caption{Values of precision, recall and F scores for Voxblox when altering minimum voxel length.}
\label{tab:voxblox-length}
\setlength{\tabcolsep}{3.5pt}
\begin{tabular}{SS*{4}{c}}
\toprule
\multicolumn{1}{M{2cm}}{Minimum Voxel Length (m)} & \multicolumn{1}{c}{Precision} & Recall & F1 Score & F2 Score & F3 Score \\ \midrule
0.05  & 0.313 & 0.043  & 0.075 & 0.052  & 0.047 \\
0.075 & 0.625 & 0.068  & 0.123 & 0.083  & 0.075 \\
0.1   & 0.749 & 0.089  & 0.159 & 0.108  & 0.097 \\
0.125 & 0.750 & 0.179  & 0.289 & 0.211  & 0.194 \\
0.150 & 0.770 & \textbf{0.550}  & 0.643 & \textbf{0.585}  & \textbf{0.568} \\
0.175 & 0.85  & 0.521  & \textbf{0.646} & 0.565  & 0.542 \\
0.2   & \textbf{0.9}   & 0.485  & 0.633 & 0.535  & 0.509 \\
\bottomrule
\end{tabular}
\end{table}

The impact of Voxblox truncation distance is elaborated upon in \autoref{tab:voxblox-truncation-distance}, providing additional support for the authors' previous argument regarding the importance of achieving a ratio of at least 1.5. 
The table clearly demonstrates that significantly better results are obtained when the ratio reaches 2, in comparison to ratios of 1 or 0.5, which is the recommended value. 
These findings emphasize the significance of appropriately adjusting the truncation distance to establish a favorable balance between voxel size and the distance of influence. 
By ensuring a higher ratio, the algorithm selectively incorporates points within close proximity to the voxel, effectively mitigating the introduction of extraneous signals originating from distant points. 
Consequently, this approach engenders more precise and dependable distance values, thereby optimizing the performance of Voxblox within the given contextual parameters.


\begin{table}[h!]
\caption{Values of precision, recall and F scores for Voxblox when altering the truncation distance.}
\label{tab:voxblox-truncation-distance}
\setlength{\tabcolsep}{3.5pt}
\begin{tabular}{S*{5}{c}}
\toprule
\multicolumn{1}{M{2cm}}{Truncation Distance (m)} & Precision & Recall & F1 Score & F2 Score & F3 Score\\
\midrule
0.05 & 0.544 & \textbf{0.698} & \textbf{0.612} & \textbf{0.661} & \textbf{0.679} \\
0.1  & \textbf{0.749} & 0.089 & 0.159 & 0.108 & 0.097 \\
0.2  & 0.634 & 0.024 & 0.045 & 0.029 & 0.026 \\
\bottomrule
\end{tabular}%
\end{table}

The impact of the Voxblox constant weight is illustrated in \autoref{tab:voxblox-constant-weight}. 
Notably, the utilization of a constant weight leads to a decline in performance compared to Voxblox's default behavior. 
This is primarily attributed to the fact that the constant weight approach overlooks crucial parameters that are typically considered by Voxblox, including the distance from the sensor to the voxel and the distance of detection to the center of the voxel. 
By neglecting these important factors, the algorithm's performance is compromised, resulting in suboptimal outcomes. 
It is evident that the dynamic and context-aware weight assignment employed by Voxblox provides more accurate and reliable results compared to the simplified constant weight approach.


\begin{table}[h!]
\caption{Values of precision, recall and F scores for Voxblox when altering if the calculated weight of a voxel in constant.}
\label{tab:voxblox-constant-weight}
\setlength{\tabcolsep}{3.5pt}
\begin{tabular}{l*{5}{c}}
\toprule
\multicolumn{1}{M{2cm}}{Constant Weight} & Precision & Recall & F1 Score & F2 Score & F3 Score\\
\midrule
\hspace{2em}True & 0.569 & 0.026 & 0.049 & 0.032 & 0.028\\
\hspace{2em}False& \textbf{0.749} & \textbf{0.089} & \textbf{0.159} & \textbf{0.108} & \textbf{0.097}\\
\bottomrule
\end{tabular}
\end{table}
\section{Conclusions}\label{sec:conclusions}
This paper presented a novel test environment designed to evaluate the performance of volumetric detection algorithms. 
The proposed environment utilizes simulation techniques to capture ground truth data and compares them to the detections obtained, enabling the calculation of precision, recall, and F scores.

To capture the ground truth data, the test environment leverages Gazebo and its LinkStates messages to track the location of each actor joint in real time. 
For each joint, a polyhedron that encompasses the entire joint is defined.
Subsequently, point in polyhedron algorithms are applied to calculate the actor points inside each joint, which are then used to construct a voxel grid correspondent to the ground truth data. 
This ground truth is compared to the voxel grid generated by the frameworks under evaluation, allowing for the determination of true positives, false positives, and false negatives.

This approach fills a gap in the field by providing a standardized method for evaluating the precision and recall of volumetric detection algorithms. 
Typically, existing approaches only compare integration time and memory usage among frameworks, neglecting other important aspects.

The development of the innovative test environment and performance metrics proved essential in providing previously unknown insights into the performance of various volumetric detection frameworks. 
The conducted tests involved systematically varying parameters within each framework to identify an optimal configuration. 
The results were evaluated using precision, recall, and F scores, supplemented by a qualitative assessment of each framework's default settings, enabling a more comprehensive analysis.

Based on these insights, it can be concluded that OctoMap, with its default parameters, is the most suitable framework for volumetric detection in a cell environment. 
While SkiMap exhibited marginally superior quantitative outcomes, the qualitative evaluation highlighted its limitation in accurately removing occupied voxels that are actually empty. 
This issue arises as a significant portion of the voxels traversed by the actor continue to be labeled as occupied in SkiMap.

Future work should focus on expanding the volume of analysis to encompass the entire trajectory of the actor during the experiment. 
This would enhance the compatibility between quantitative and qualitative analyses and address the aforementioned issue with SkiMap. 
Additionally, further tests of the test environment should be conducted in different scenarios and with a wider range of frameworks to ensure comprehensive evaluation and validation.


\printbibliography

\begin{IEEEbiography}[{\includegraphics[width=1in,height=1.25in,clip,keepaspectratio]{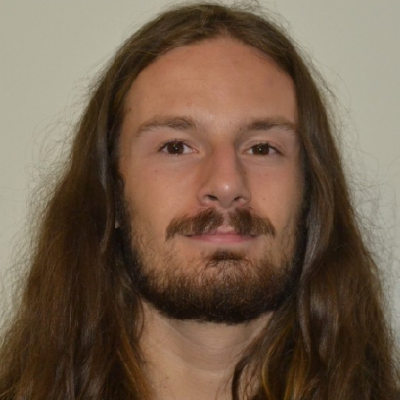}}]{Manuel Gomes} received an Integrated Master's Degree in Mechanical Engineering 
from the University of Aveiro in 2022.
Their M.Sc. dissertation focused on sensor calibration in robotic vehicles. 
Currently pursuing a Ph.D. in Mechanical Engineering at the University of Aveiro, the author is actively involved in the AUGMANITY project, a EU project, working on collaborative robots. 
\end{IEEEbiography}

\begin{IEEEbiography}[{\includegraphics[width=1in,height=1.25in,clip,keepaspectratio]{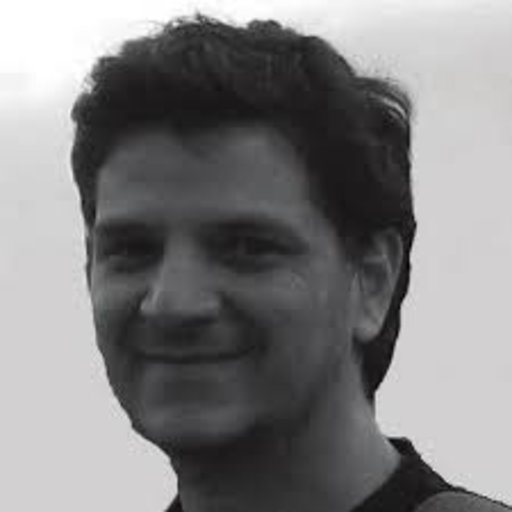}}]{Miguel Oliveira} 
received the bachelor's and
M.Sc. degrees from the University of Aveiro,
Aveiro, Portugal, in 2004 and 2007, respectively, and the Ph.D. degree in mechanical
engineering (specialization in robotics, on the
topic of autonomous driving systems), in 2013.
From 2013 to 2017, he was an Active Researcher
with the Institute of Electronics and Telematics Engineering of Aveiro, Aveiro, and with the
Institute for Systems and Computer Engineering,
Technology and Science, Porto, Portugal, where he participated in several
EU-funded projects, as well as national projects. He is currently an Assistant
Professor with the Department of Mechanical Engineering, University of
Aveiro, where he was also the Director of the Masters in Automation Engineering. 
He has supervised more than 20 M.Sc. students. He authored over
20 journal publications, from 2015 to 2020. His research interests include
autonomous driving, visual object recognition in open-ended domains, multi-modal sensor fusion, computer vision, and the calibration of robotic systems.
\end{IEEEbiography}

\begin{IEEEbiography}[{\includegraphics[width=1in,height=1.25in,clip,keepaspectratio]{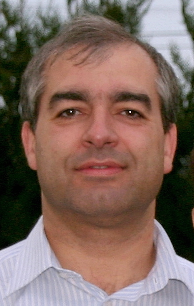}}]{Vítor Santos} 
obtained a 5-year degree in Electronics Engineering and Telecommunications in 1989, at the University of Aveiro, Portugal, where he later obtained a PhD in Electrical Engineering in 1995, and the Habilitation in Mechanical Engineering in 2018. 
He was awarded fellowships for research in mobile robotics during 1990-1994 at the Joint Research Center, Italy. 
He is currently Associate Professor at the University of Aveiro where he lectures several courses on robotics, autonomous vehicles and computer vision, and has carried out research activity on mobile robotics, autonomous driving, advanced perception and humanoid robotics, also in public and privately funded projects. 
He has supervised and co-supervised more than 100 students in Masters, PhD and Postdoc programs, and coordinated the creation of two University degrees in the field of Automation at the University of Aveiro. 
He founded the ATLAS project for mobile robot competition that achieved 6 first prizes in the annual Autonomous Driving competition, and has coordinated the development of ATLASCAR, the first real car with autonomous navigation capabilities in Portugal. 
He was involved in the organization of several conferences, workshops and special sessions in national and international events, including being the General Chair of the IEEE International Conference on Autonomous Robot Systems and Competitions, ICARSC2021. 
He is one of the founders of the Portuguese Robotics Open in 2001 and co-founder of the Portuguese Society of Robotics in 2006.

\end{IEEEbiography}

\EOD

\end{document}